\documentclass[10pt,twocolumn,letterpaper]{article}

\usepackage{cvpr}
\usepackage{times}
\usepackage{epsfig}
\usepackage{graphicx}
\usepackage{amsmath}
\usepackage{amssymb}
\usepackage{subfigure}
\usepackage{mathrsfs}
\usepackage{algorithm}
\usepackage{algorithmic}
\usepackage{multirow}
\usepackage{bm}
\usepackage{amsfonts}       % blackboard math symbols

\usepackage{color}
\usepackage[normalem]{ulem}
\usepackage{comment}

\newcommand\XY[1]{\textcolor{black}{#1}}

\newcommand\CXY[1]{\textcolor{black}{#1}}

\newcommand{\calD}{{\cal D}}

\newcommand{\calF}{{\cal F}}

\newcommand{\calR}{{\cal R}}

\newcommand{\vf}{{\bf f}}

\newcommand{\vm}{{\bf m}}

\newcommand{\vx}{{\bf x}}

\newcommand{\vz}{{\bf z}}

\newcommand{\vLambda}{{\bf{\Lambda}}}

\def\onedot{. }
\cvprfinalcopy % *** Uncomment this line for the final submission

\def\cvprPaperID{1239} % *** Enter the CVPR Paper ID here

\begin{document}

%%%%%%%%% TITLE
\title{Few-Shot Class-Incremental Learning}

\author{Xiaoyu Tao\textsuperscript{\rm 1}, Xiaopeng Hong\textsuperscript{\rm 1,3}{\thanks{Corresponding author}}, Xinyuan Chang\textsuperscript{\rm 2}, Songlin Dong\textsuperscript{\rm1}, Xing Wei\textsuperscript{\rm 2}, Yihong Gong\textsuperscript{\rm 2}\\
\textsuperscript{\rm 1}Faculty of Electronic and Information Engineering, Xi’an Jiaotong University\\
\textsuperscript{\rm 2}School of Software Engineering, Xi'an Jiaotong University\\
\textsuperscript{\rm 3}Research Center for Artificial Intelligence, Peng Cheng Laboratory\\
{\tt\small txy666793@stu.xjtu.edu.cn, hongxiaopeng@mail.xjtu.edu.cn, cxy19960919@stu.xjtu.edu.cn,} \\
{\tt\small dsl972731417@stu.xjtu.edu.cn, xingxjtu@gmail.com, ygong@mail.xjtu.edu.cn}
%\and
%Second author\\
}

\maketitle
\thispagestyle{empty}

%%%%%%%%% ABSTRACT
\begin{abstract}
   The ability to incrementally learn new classes is crucial to the development of real-world artificial intelligence systems. In this paper, we focus on a challenging but practical \emph{few-shot class-incremental learning} (FSCIL) problem. FSCIL requires CNN models to incrementally learn new classes from very few labelled samples, without forgetting the previously learned ones. To address this problem, we represent the knowledge using a \emph{neural gas} (NG) network, which can learn and preserve the topology of the feature manifold formed by different classes.  On this basis, we propose the \emph{TOpology-Preserving knowledge InCrementer} (TOPIC) framework. TOPIC mitigates the forgetting of the old classes by stabilizing NG's topology and improves the representation learning for few-shot new classes by growing and adapting NG to new training samples. Comprehensive experimental results demonstrate that our proposed method significantly outperforms other state-of-the-art class-incremental learning methods on CIFAR100, miniImageNet, and CUB200 datasets. 
\end{abstract}

%%%%%%%%% BODY TEXT
\section{Introduction}

Convolutional Neural Networks (CNNs) have been successfully applied to a broad range of computer vision tasks \cite{AlexNet,ResNet,RPN,DeepLabV3,Sphereface,CrowdCount,FGVC,reid}. For practical use, we train CNN models on large scale image datasets \cite{ImageNet} and then deploy them on smart agents. As the smart agents are often exposed in a new and dynamic environment, there is an urgent need to continuously adapt the models to recognize new classes emerging.  
For example, the smart album function on smartphones is designed to automatically classify user photos into both the pre-defined and user-defined classes. The model underpinning the smart album is pre-trained on the training set of the \emph{pre-defined} classes, and is required to adapt to the new \emph{user-defined} classes by learning from new photos. From the users' perspective, they are only willing to annotate very few image examples for the new class, as the labeling process consumes manpower. Therefore, it is crucial for CNNs to be capable of incrementally learning new classes from very few training examples. We term this ability as \emph{few-shot class-incremental learning} (FSCIL).

\begin{figure}[t]
\begin{center}
    \subfigure[]{
        \label{fig:loss:a}
        \includegraphics[width=0.23\textwidth]{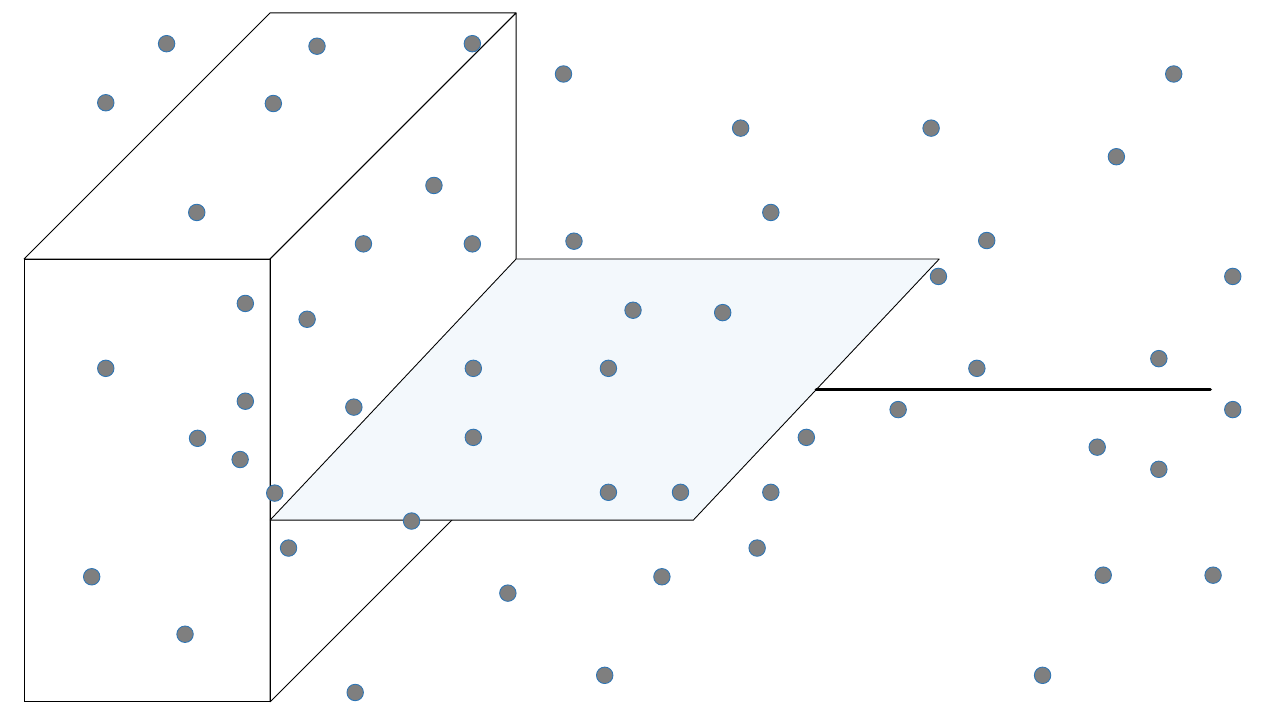}}
    \subfigure[]{
        \label{fig:loss:b}
        \includegraphics[width=0.23\textwidth]{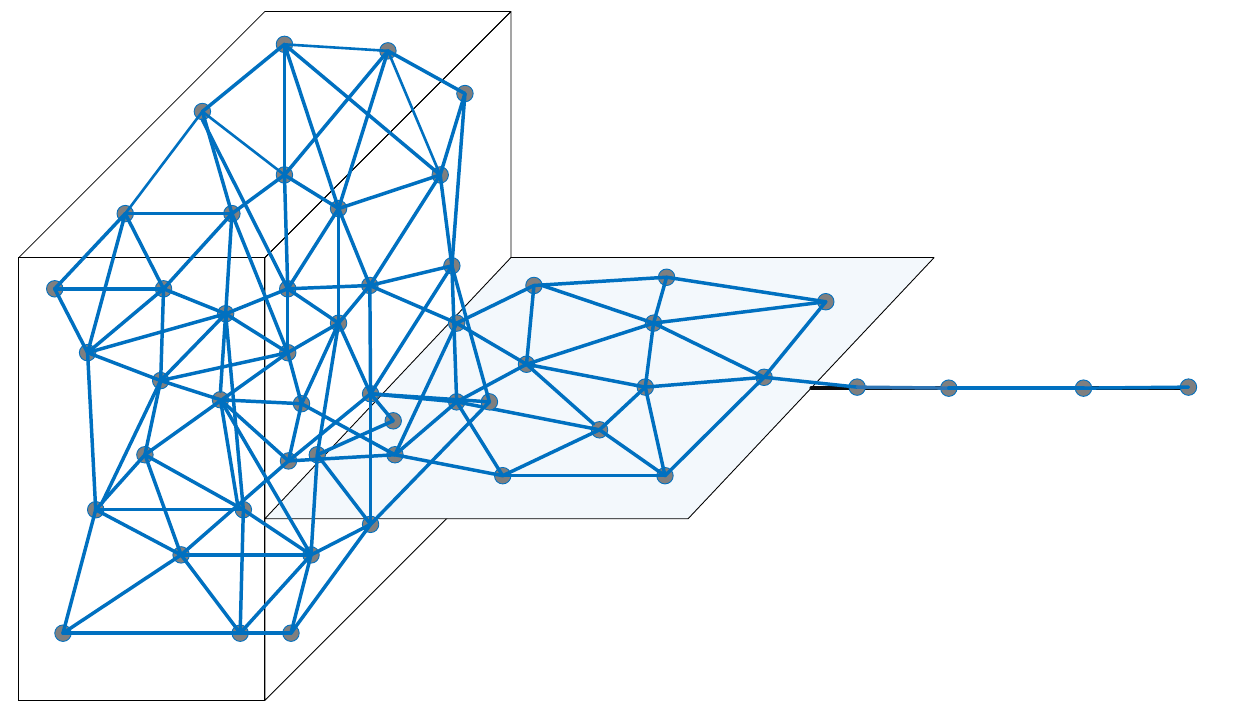}}
\end{center}
\caption{\textbf{Comparisons of two ways to characterize a heterogenous manifold.} (a) Randomly sampled representatives, which are adopted by conventional CIL studies for knowledge distillation. (b) The representatives learned by \emph{neural gas}, which well preserves the topology of the manifold.}
\label{fig:manifold}
\end{figure}

A na\"ive approach for FSCIL is to finetune the base model on the new class training set. However, a simple finetuning with limited number of training samples would cause two severe problems: one is ``forgetting old'', where the model's performance deteriorates drastically on old classes due to \emph{catastrophic forgetting}~\cite{CF}; the other is ``overfitting new'', where the model is prone to overfit to new classes, which loses generalization ability on large set of test samples. 

Recently, there have been many research efforts attempting to solve the \emph{catastrophic forgetting} problem~\cite{EWC,SI,LWF,GEM,IMM,ICARL,EEIL,NCM,BOCL,DGR,MAS}. They usually conduct incremental learning under the multi-task or the multi-class scenarios. The former incrementally learns a sequence of disjoint tasks, which requires the task identity in advance. This is seldom satisfied in real applications where the task identity is typically unavailable. The latter learns a unified classifier to recognize all the encountered classes within a single task. This scenario is more practical without the need of knowing task information. In this paper, we study the FSCIL problem under the multi-class scenario, where we treat FSCIL as a particular case of the \emph{class-incremental learning} (CIL) \cite{ICARL,EEIL,EGR,NCM,BIC}. Compared with CIL that learns new classes with unlimited, usually large-scale training samples, FSCIL is more challenging, since the number of new training samples is very limited.
 
%Recent works in \emph{class-incremental learning} (CIL) \cite{ICARL,EEIL,PDR,NCM,BIC} typically adopt the multi-class scenario.

To mitigate forgetting, most CIL works \cite{ICARL,EEIL,PDR,NCM,BIC} use the \emph{knowledge distillation} \cite{KD} technique that maintains the network's output logits corresponding to old classes. They usually store a set of old class exemplars and apply the distillation loss to the network's output. Despite their effectiveness, there are several problems when training with the distillation loss. One is the \emph{class-imbalance} problem \cite{NCM,BIC}, where the output logits are biased towards those classes with a significant larger number of training samples. The other is the \emph{performance trade-off}  between old and new classes. This problem is more prominent for FSCIL, because learning from very few training samples requires a larger learning rate and stronger gradients from new classes' classification loss, making it difficult to maintain the output for old classes at the same time.

In this paper, we address FSCIL from a new, cognitive-inspired perspective of knowledge representation.
Recent discoveries in cognitive science reveal the importance of topology preservation for maintaining the memory of the old knowledge~\cite{topology, topology2}. The change of the memory's topology will cause severe degradation of human recognition performance on historical visual stimuli~\cite{topology}, indicating \emph{catastrophic forgetting}. Inspired by this, we propose a new FSCIL framework, named \emph{TOpology-Preserving knowledge InCrementer} (TOPIC),  as shown in Figure~\ref{fig:manifold}. TOPIC uses a \emph{neural gas} (NG) network~\cite{NG,GNG,IGNG} to model the topology of feature space. When learning the new classes, NG grows to adapt to the change of feature space. On this basis, we formulate FSCIL as an optimization problem with two objectives. On the one hand, to avoid \emph{catastrophic forgetting}, TOPIC preserves the old knowledge by stabilizing the topology of NG, which is implemented with an \emph{anchor loss} (AL) term. On the other hand, to prevent overfitting to few-shot new classes, TOPIC adapt the feature space by pushing the new class training sample towards a correct new NG node with the same label and pulling the new nodes of different labels away from each other. The \emph{min-max loss} (MML) term is developed to achieve this purpose.

For extensive assessment, we build the FSCIL baselines by adapting the state-of-the-art CIL methods \cite{ICARL,EEIL,NCM} to this new problem and compare our method with them. We conduct comprehensive experiments
%\footnote{We introduce a new  \emph{aquisition-memory} (AM) metric to measure the trade-off between old and new classes in Supplementary Material.} 
on the popular CIFAR100 \cite{cifar}, miniImageNet \cite{miniImageNet}, and CUB200~\cite{cub200} datasets.  Experimental results demonstrate the effectiveness of the proposed FSCIL framework. 

To summarize, our main contributions include:
\begin{itemize}
\item We recognize the importance of \emph{few-shot class-incremental learning} (FSCIL) and define a problem setting to better organize the FSCIL research study. Compared with the popularly studied \emph{class-incremental learning} (CIL), FSCIL is more challenging but more practical. 
\item We propose an FSCIL framework TOPIC that uses a \emph{neural gas} (NG) network to learn feature space topologies for knowledge representation. TOPIC stabilizes the topology of NG for mitigating forgetting and adapts NG to enhancing the discriminative power of the learned features for few-shot new classes.
\item We provide an extensive assessment of the FSCIL methods, which we adapt the state-of-the-art CIL methods to FSCIL and make comprehensive comparisons with them.
\end{itemize}

\section{Related Work}

\subsection{Class-Incremental Learning}

\emph{Class-incremental learning} (CIL) learns a unified classifier incrementally to recognize all encountered new classes met so far. To mitigate the forgetting of the old classes, CIL studies typically adopt the knowledge distillation technique, where external memory is often used for storing old class exemplars to compute the distillation loss. 
For example, iCaRL \cite{ICARL} maintains an ``episodic memory'' of the exemplars and incrementally learns the nearest-neighbor classifier for the new classes. EEIL \cite{EEIL} adds the distillation loss term to the cross-entropy loss for end-to-end training. Latest CIL works NCM~\cite{NCM} and BiC~\cite{BIC} reveal the class-imbalance problem that causes the network's prediction biased towards new classes. They adopt cosine distance metric to eliminate the bias in the output layer \cite{NCM}, or learns a bias-correction model to post-process the output logits \cite{BIC}. 
%In this way, CIL can be scaled up to a large number of classes. 

In contrast to these CIL works, we focus on the more difficult FSCIL problem, where the number of new class training samples is limited. Rather than constraining the network's output, we try to constrain CNN's feature space represented by a neural gas network. 
%In this way, we can avoid the issues of the distillation approaches and achieve better effect for mitigating forgetting and improving the learning on few-shot examples. 

\subsection{Multi-task Incremental Learning}

A series of research works adopts the multi-task incremental learning scenario. These works can be categorized into three types: (1) rehearsal approaches~\cite{GEM,AGEM,DGR,LGAN,MGR}, (2) architectural approaches~\cite{PackNet,Piggyback,MAS,HAT,DEN}, and (3) regularization approaches~\cite{EWC,SI,REWC,IMM}.
Rehearsal approaches replay the old tasks information to the task solver when learning the new task. One way is to store the old tasks' exemplars using external memory and constrain their losses during learning the new task~\cite{GEM,AGEM}. Another way is to use the generative models to memorize the old tasks’ data distribution~\cite{DGR,MGR,LGAN}. For example, DGR~\cite{DGR} learns a generative adversarial network to produce observed samples for the task solver. The recognition performance is affected by the quality of the generated samples. Architectural approaches alleviate forgetting by manipulating the network's architecture, such as network pruning, dynamic expansion, and parameter masking. For example, PackNet~\cite{PackNet} prunes the network to create free parameters for the new task. HAT~\cite{HAT} learns the attention masks for old tasks and use them to constrain the parameters when learning the new task. Regularization approaches impose regularization on the network's parameters, losses or output logits. 
For example, EWC~\cite{EWC} and its variants~\cite{SI,REWC} penalize the changing of the parameters important to old tasks. These methods are typically based on certain assumptions of the parameters' posterior distribution (e.g. Gaussian), which may struggle in more complex scenarios. 

As the multi-task incremental learning methods are aimed at learning disjoint tasks, it is infeasible to apply these methods under the single-task multi-class scenario adopted by FSCIL. As a result, we have to exclude them for comparison.

\subsection{Dynamic Few-Shot Learning}

\emph{Few-shot learning} (FSL) aims to adapt the model to recognize unseen novel classes using very few training samples, while the model's recognition performance on the base classes is not considered. To achieve FSL, research studies usually adopt the metric learning and meta-learning strategies~\cite{miniImageNet,PN,RN,MAML,MTL}.
Recently, some FSL research works attempt to learn a model capable of recognizing both the base and novel classes \cite{DFS,IFS}. Typically, they first pretrain the model on the base training set to learn feature embedding as well as the weights of the classifier for base classes. Then they perform meta-learning for few-shot novel classes, by sampling ``fake'' few-shot classification tasks from the base dataset to learn a classifier for novel classes. Finally, the learned heads are combined for recognizing the joint test (query) set of the base and novel classes. 

Though some of these works \cite{IFS} regard such setting as a kind of incremental learning, they rely on the old training set (i.e., the base class dataset) for sampling meta-learning tasks. This is entirely different from the FSCIL setting, where the base/old class training set is unavailable at the new incremental stage. As a consequence, these few-shot learning works can not be directly applied to FSCIL.

\section{Few-Shot Class-Incremental Learning}

We define the \emph{few-shot class-incremental-learning} (FSCIL) setting as follows. Suppose we have a stream of labelled training sets $D^{(1)}, D^{(2)}, \cdots$, where $D^{(t)} = \lbrace (\vx_j^{(t)}, y_j^{(t)}) \rbrace_{j=1}^{\vert D^{(t)} \vert}$. $L^{(t)}$ is the set of classes of the $t$-th training set, where $\forall i,j,\ L^{(i)} \cap L^{(j)}=\varnothing$. $D^{(1)}$ is the large-scale training set of base classes, and $D^{(t)}, t>1$ is the few-shot training set of new classes.  The model $\Theta$ is incrementally trained on $D^{(1)}, D^{(2)}, \cdots$ with a \emph{unified} classification layer, while only $D^{(t)}$ is available at the $t$-th training session. After training on $D^{(t)}$, $\Theta$ is tested to recognize all encountered classes in $L^{(1)}, \cdots, L^{(t)}$. For $D^{(t)}, t>1$, we denote the setting with $C$ classes and $K$ training samples per class as the \emph{C-way K-shot} FSCIL.
The main challenges are two-fold: (1) avoiding \emph{catastrophic forgetting} of old classes; (2) preventing overfitting to few-shot new classes.

%\subsection{Preliminary}
To perform FSCIL, we treat the CNN as a composition of a feature extractor $f(\cdot;\theta)$ with the parameter set $\theta$  and a classification head. The feature extractor defines the feature space $\calF \subseteq \mathbb{R}^n$. The classification head with the parameter set $\phi$ produces the output vector followed by a softmax function to predict the probability $p$ over all classes. The entire set of parameters is denoted as $\Theta=\lbrace \theta, \phi \rbrace$. The output vector given input $\vx$ is $o(\vx; \Theta) = \phi^T f(\vx; \theta)$. 
Initially, we train $\Theta^{(1)}$ on $D^{(1)}$ with the cross-entropy loss. Then we incrementally finetune the model on $D^{(2)},D^{(3)},\cdots$, and get $\Theta^{(2)},\Theta^{(3)},\cdots$. 
At the $t$-th session ($t>1$), the output layer is expanded for new classes by adding $\vert L^{(t)} \vert$ output neurons. 

For FSCIL, we first introduce a baseline solution to alleviate forgetting based on knowledge distillation; then we elaborate our proposed TOPIC framework that employs a neural gas network for knowledge representation and the \emph{anchor loss} and \emph{min-max loss} terms for optimization.

\subsection{Baseline: Knowledge Distillation Approach}\label{sec:baseline}

Most CIL works \cite{ICARL,EEIL,NCM,BIC} adopt the \emph{knowledge distillation} technique for mitigating forgetting. Omitting the superscript $(t)$, the loss function is defined as:
\begin{equation}\label{eq:dist}
\ell(D, P; \Theta) = \ell_{CE} (D, P; \Theta) + \gamma \ell_{DL} (D, P; \Theta),
\end{equation}
where $\ell_{DL}$ and $\ell_{CE}$ are the distillation and cross-entropy loss terms, and $P$ is the set of old class exemplars drawn from $D^{(1)},\cdots,D^{(t-1)}$.
The implementation of $\ell_{DL}$ may vary in different works. Generally, it takes the form:
\begin{align}
\ell_{DL} (D, P; \Theta) &= \sum_{(\vx,y) \in  D \cup P} \sum_{k=1}^{n} - \tau_k(\vx; \hat{\Theta}) \log(\tau_k(\vx;\Theta)), \nonumber \\
\tau_k(\vx; \Theta) &= \frac{e^{o_k(\vx; \Theta)/T}}{\sum_{j=1}^{n} e^{o_j(\vx; \Theta)/T}},
\end{align}
where $n = \sum_{i=1}^{t-1} \vert L^{(i)} \vert$ is the number of the old classes, $\hat{\Theta}$ is the initial values of $\Theta$ before finetuning, and $T$ is the distillation temperature (e.g., $T=2$ in~\cite{EEIL,NCM}).

The distillation approach faces several critical issues when applied to FSCIL. One is the bias problem caused by imbalanced old/new class training data, where the output layer is biased towards new classes~\cite{NCM,BIC}. To address this issue, \cite{NCM} uses cosine distance measure to eliminate the bias and \cite{BIC} learns a bias correction model to post-process the outputs. Despite their effectiveness in learning large-scale training data, they are less effective for FSCIL with very few training samples. Using cosine distance may lose important patterns (e.g. appearance) contained in the magnitude of the weight/feature vector, while the bias-correction model requires a large number of training samples, which conflicts with the few-shot setting. 
Another issue is the dilemma to balance the contribution between $\ell_{CE}$ and $\ell_{DL}$, which may lead to unsatisfactory performance trade-off. Learning few-shot new classes requires a larger learning rate to minimize $\ell_{CE}$, while it can cause instability of the output logits and makes it difficult to minimize $\ell_{DL}$. 

Based on the above considerations, we abandon the distillation loss in our framework. Instead, we manipulate the knowledge contained CNN's feature space that contains richer information than the output logits.

\subsection{Knowledge Representation as Neural Gas}\label{sec:ng}

\begin{figure}[t]
\begin{center}
    \includegraphics[width=0.44\textwidth]{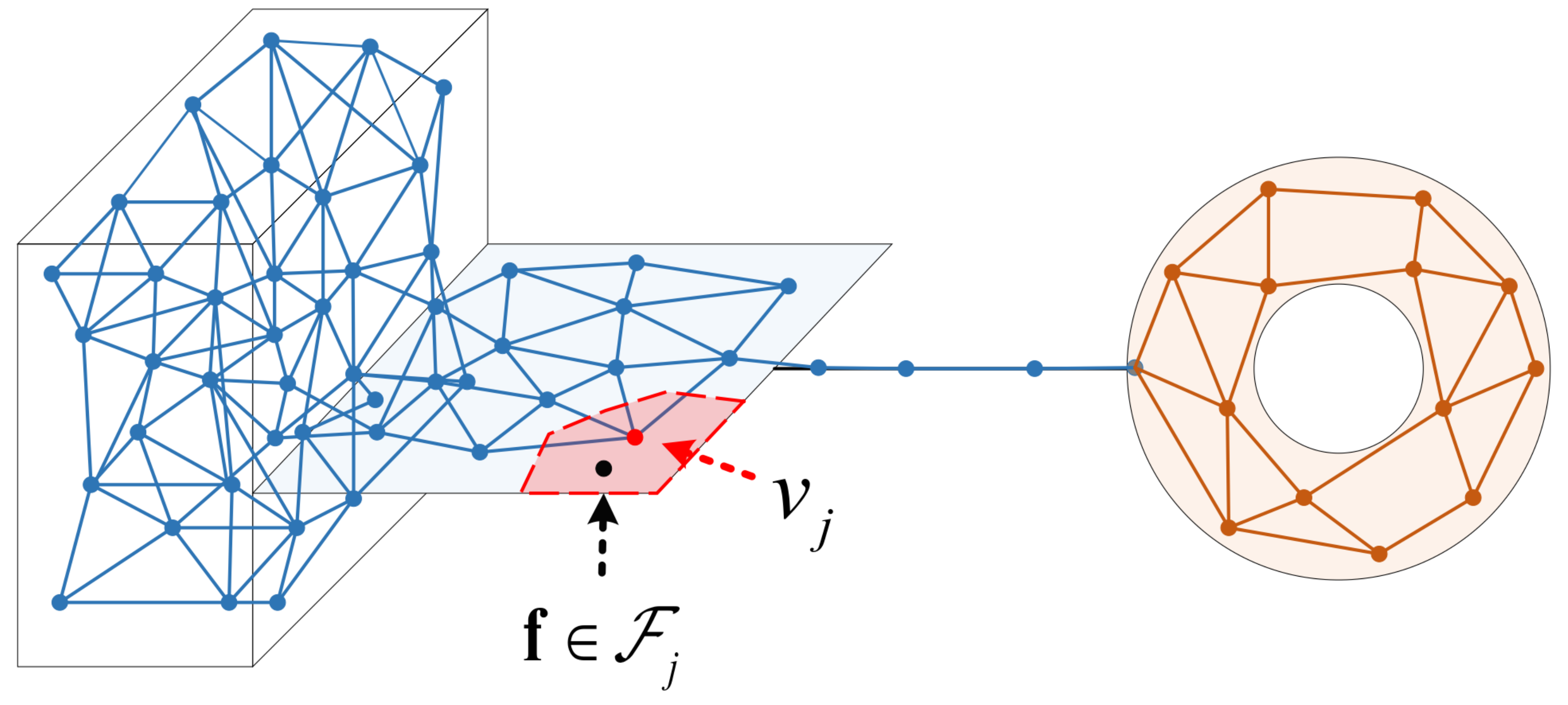}
\end{center}
\caption{\textbf{NG preserves the topology of heterogenous feature space manifold.} Initially, NG is learnt for base classes (the blue dots and lines.) Then NG incrementally grows for new classes by inserting new nodes and edges (the orange dots and lines.)  During the competitive Hebbian learning, $v_j$'s centroid vector $\vm_j$ is adapted to the input vector $\vf$ which falls in  $\calF_j$ encoded by $v_j$. }
\label{fig:ngnet}
\end{figure}

The knowledge distillation methods typically store a set of exemplars randomly drawn from the old training set and compute the distillation loss using these exemplars. However, there is no guarantee that the randomly-sampled exemplars can well represent heterogenous, non-uniform data of different classes in the FSCIL scenarios.
%the randomly sampled exemplars \XY{can not well represent heterogenous feature space trained with non-uniform data of different classes, which makes it less capable of preserving old knowledge.}
Instead, we represent the knowledge by preserving the feature space topology, which is achieved by a \emph{neural gas} (NG) network~\cite{NG}.  
NG maps the feature space $\calF$ to a finite set of feature vectors $V=\lbrace v_j \rbrace_{j=1}^N$ and preserves the topology of $\calF$ by \emph{competitive Hebbian learning}~\cite{CHL}, as shown in Figure~\ref{fig:ngnet}. 

NG defines an undirected graph $G=\langle V, E \rangle$. Each vertex $v_j \in V$ is assigned with a centroid vector $\vm_j \in \mathbb{R}^n$ describing the location of $v_j$ in feature space. The edge set $E$ stores the neighborhood relations of the vertices. If $v_i$ and $v_j$ are topologically adjacent, $e_{ij}=1$; otherwise, $e_{ij}=0$. Each edge $e_{ij}$ is assigned with an ``age'' $a_{ij}$ initialized to 0. Given an input $\vf \in \calF$, it matches the NG node $j$ with the minimum distance $d(\vf, \vm_j)$ to $\vf$.
%$j = \mathop{\arg \min}_{i} \ d(\vf, \vm_i) \ \forall i \in \lbrace 1,\cdots,N\rbrace$.
The matching process divides $\calF$ into disjoint subregions, where the centroid vector $\vm_j$ encodes the region $\calF_j = \lbrace \vf \in \calF \vert d(\vf, \vm_j) \leq d(\vf, \vm_i), \ \forall i \rbrace$. We use the \emph{Euclidean distance} as $d(\cdot, \cdot)$.

\begin{figure*}[t]
\begin{center}
    \subfigure[]{
        \label{fig:loss:a}
        \includegraphics[width=0.15\textwidth]{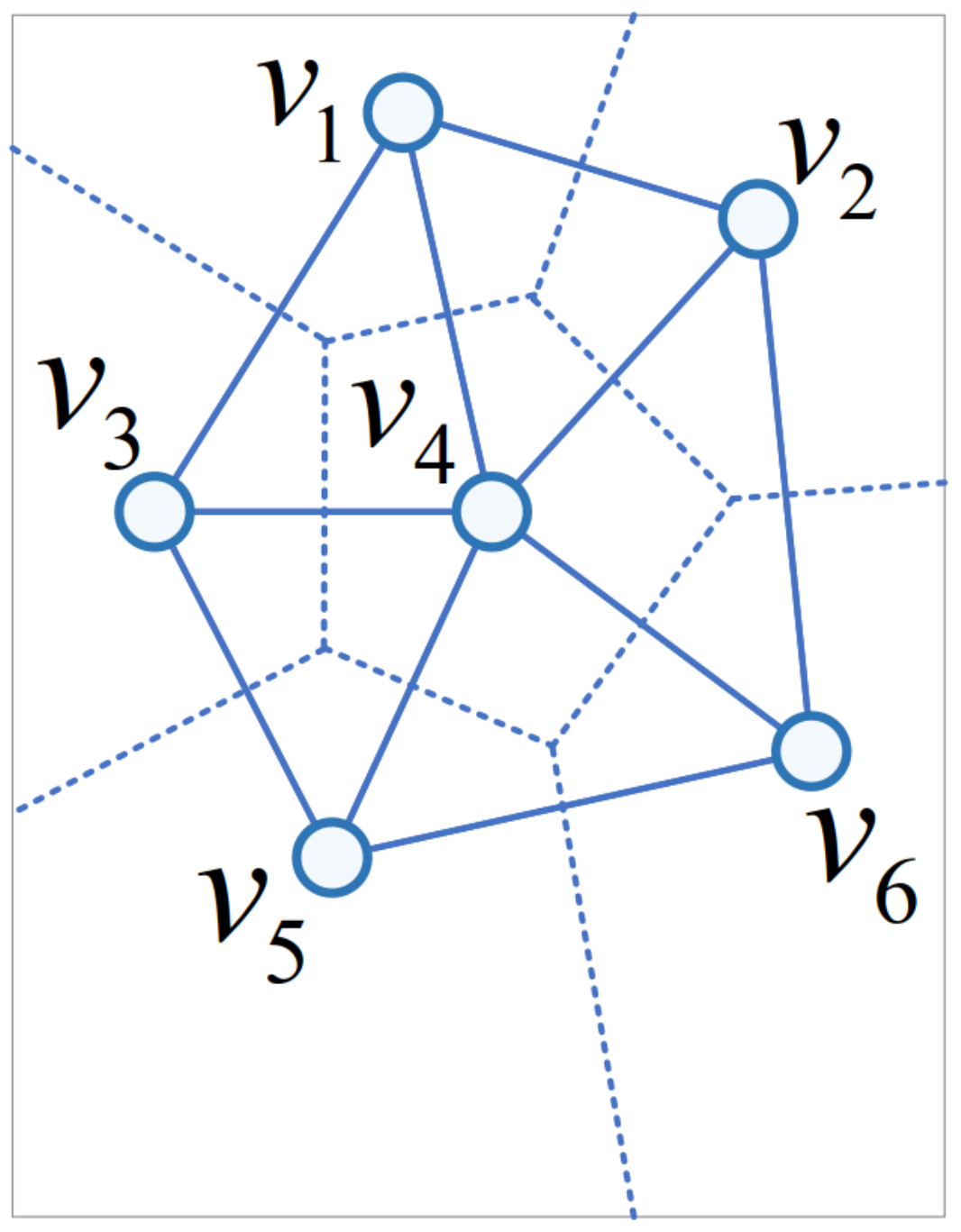}}
    \subfigure[]{
        \label{fig:loss:b}
        \includegraphics[width=0.15\textwidth]{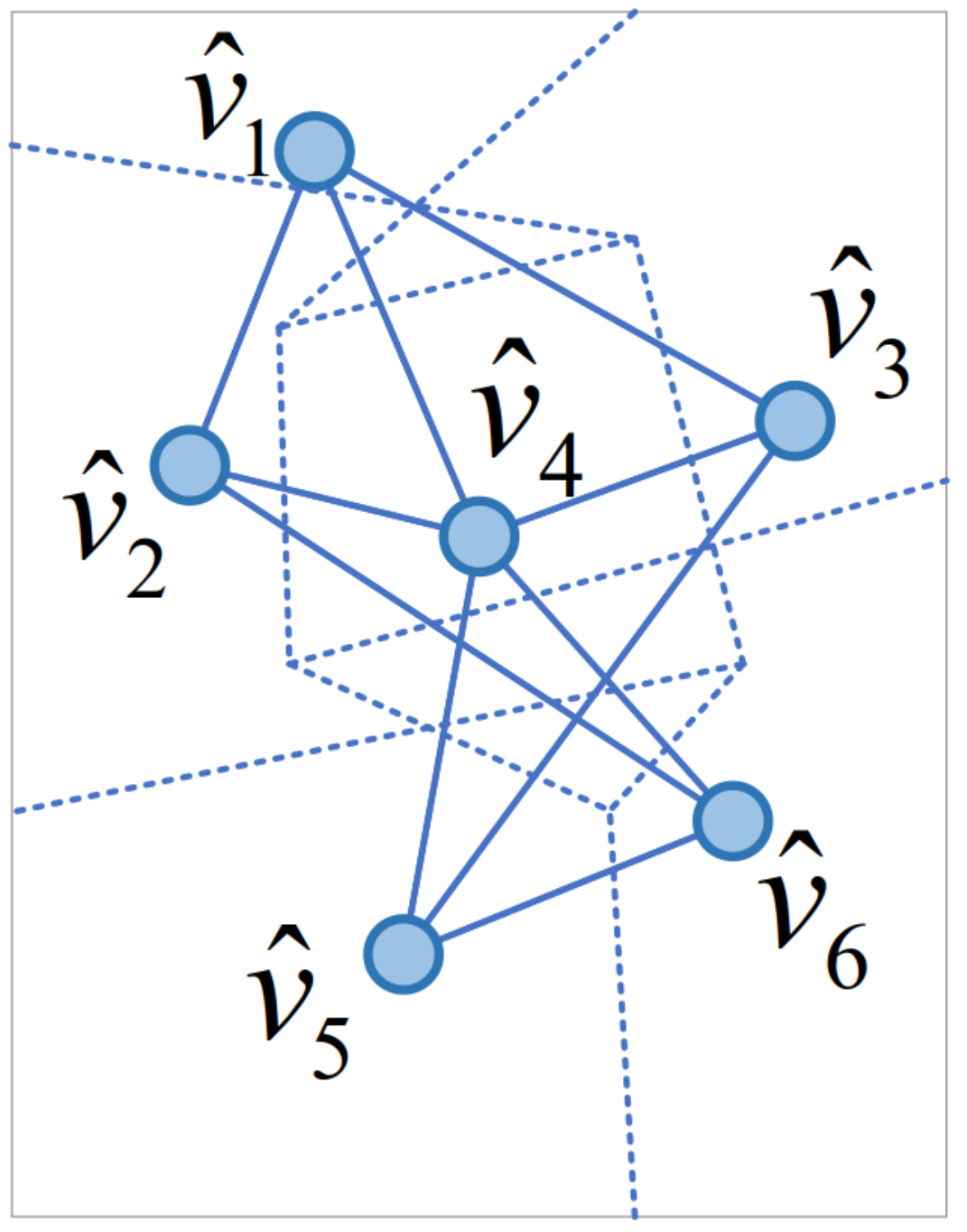}}
    \subfigure[]{
        \label{fig:loss:c}
        \includegraphics[width=0.15\textwidth]{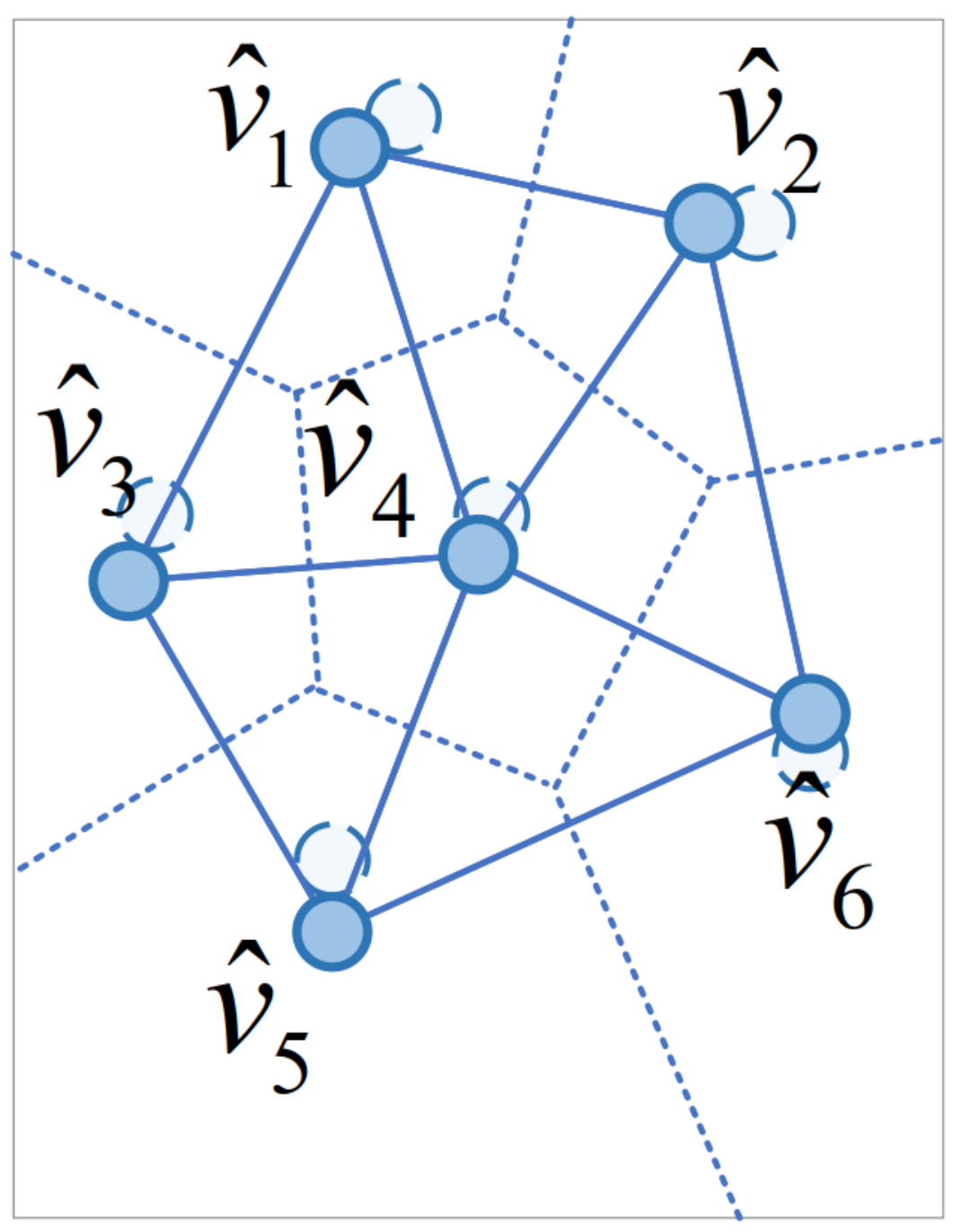}}
    \subfigure[]{
        \label{fig:loss:d}
        \includegraphics[width=0.15\textwidth]{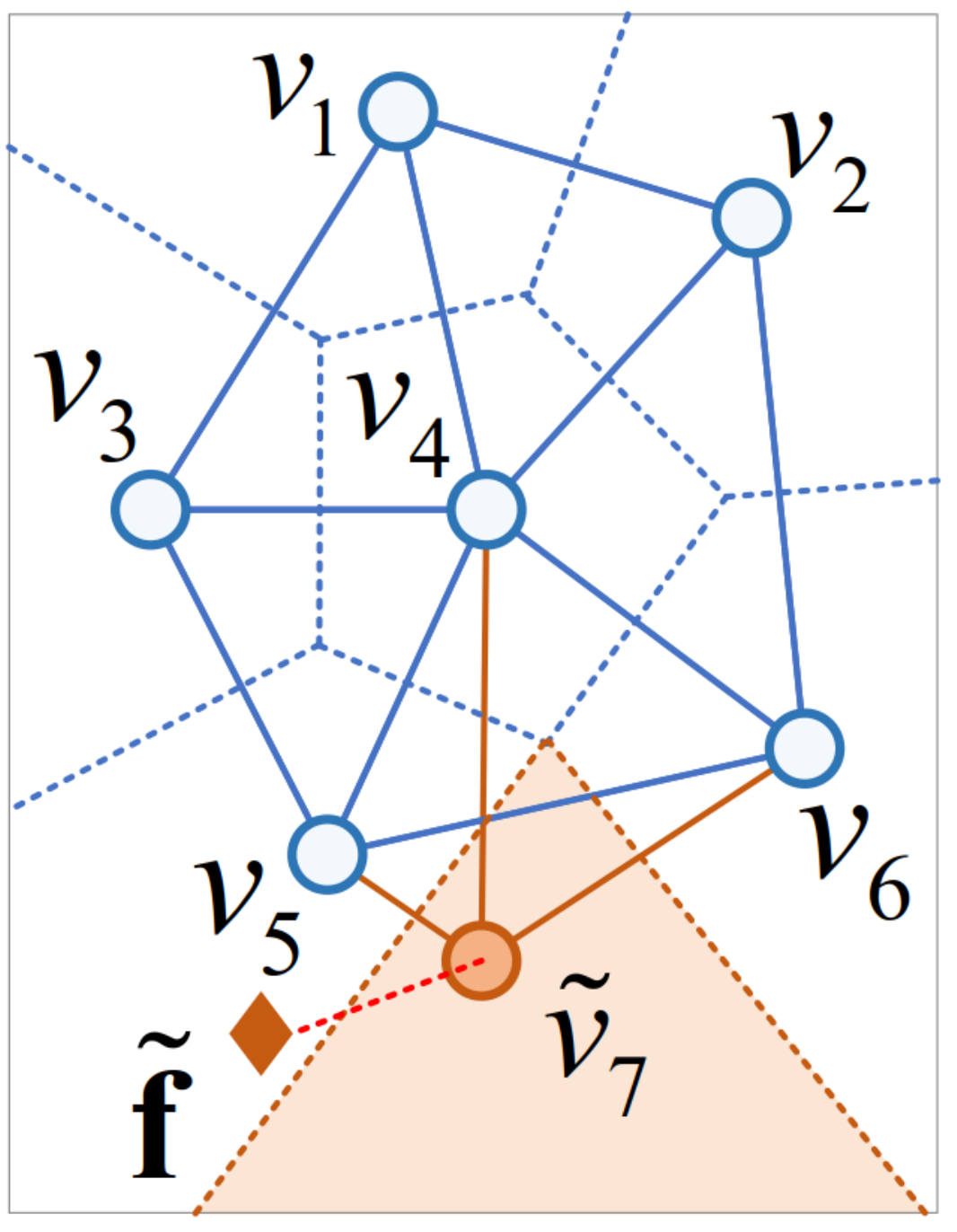}}
    \subfigure[]{
        \label{fig:loss:e}
        \includegraphics[width=0.15\textwidth]{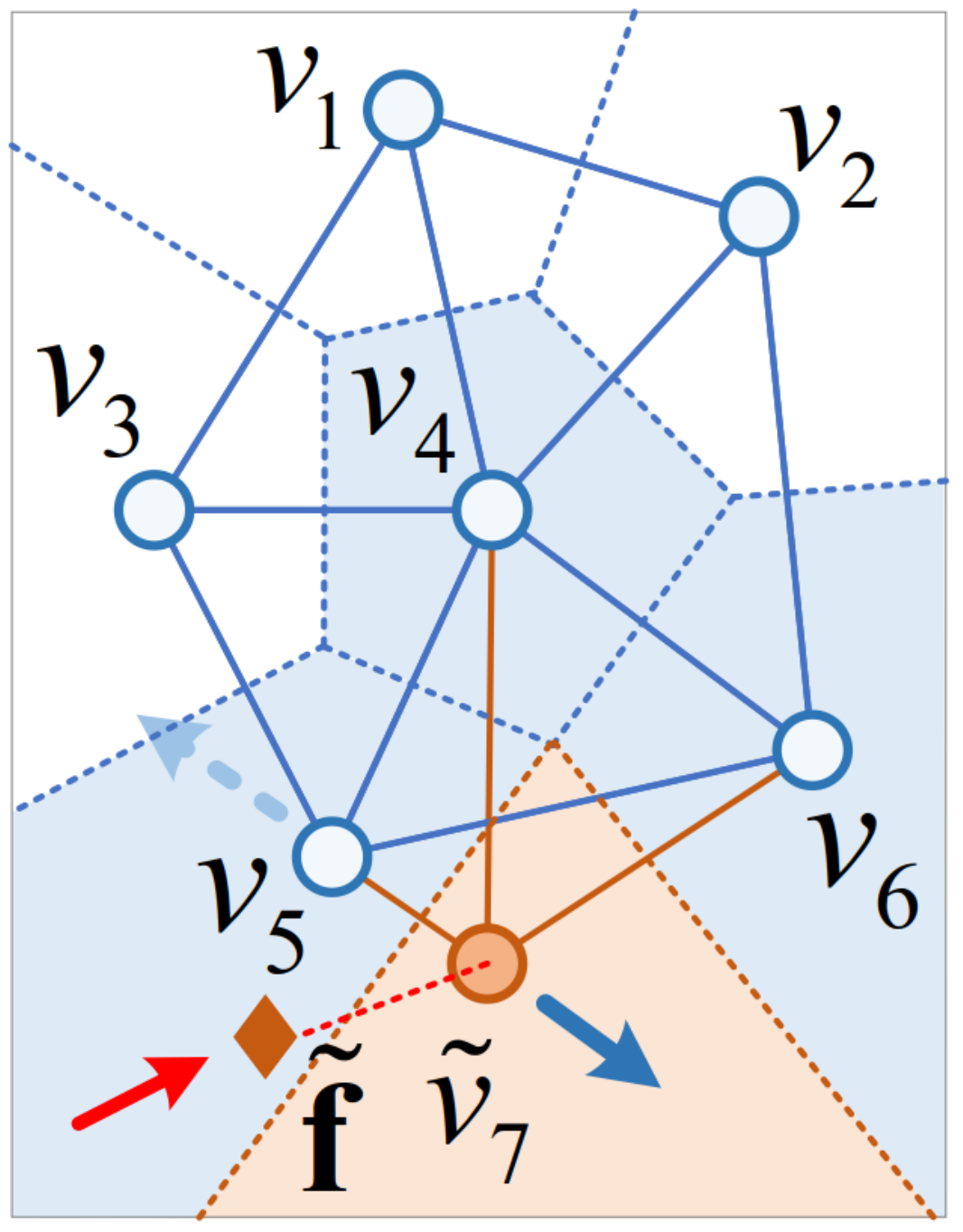}}
    \subfigure[]{
        \label{fig:loss:f}
        \includegraphics[width=0.15\textwidth]{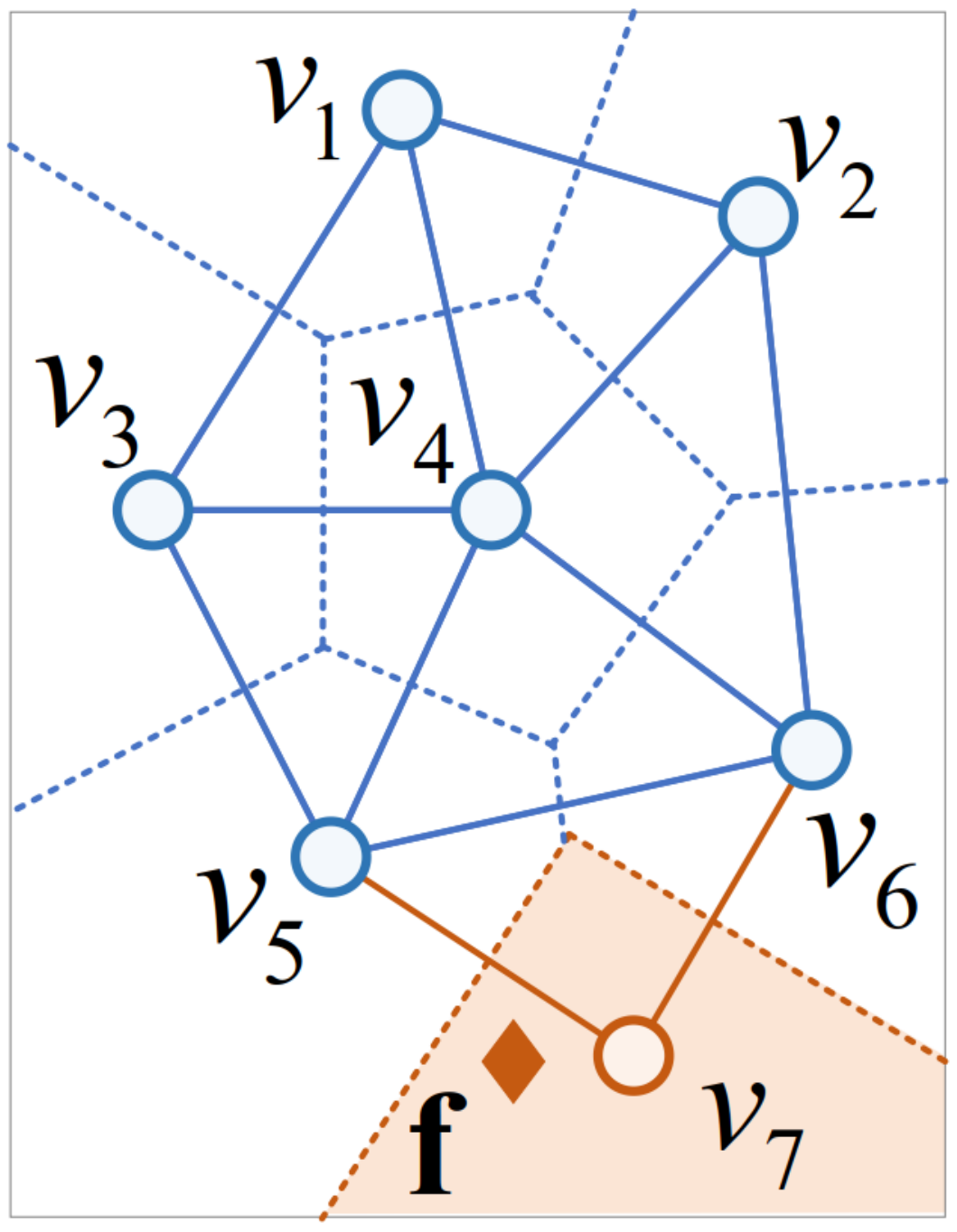}}
\end{center}
\caption{\textbf{Explanation of NG stabilization and adaptation.} (a) NG divides CNN's feature space $\calF$ into a set of topologically arranged subregions $\calF_j$ represented by a centroid vector $v_j$. (b) When finetuning CNN with few training examples, $\calF$'s topology is severely distorted, indicating \emph{catastrophic forgetting}. (c) To maintain the topology, the shift of NG nodes is penalized by the \emph{anchor-loss} term. (d) NG grows for new class $y$ by inserting a new vertex $\tilde{v}_7$. A new class training sample $\tilde{\vf}$ is mismatched to $v_5$, due to $d(\tilde{\vf}, \vm_5) < d(\tilde{\vf}, \vm_7)$. (e) The \emph{min-max loss} term adapts $\calF_7$ by pushing $\tilde{\vf}$ to $\tilde{v}_7$ and pulling $\tilde{v}_7$ away from the neighbors $v_4$, $v_5$ and $v_6$. (f) The topology is updated after the adaptation in (e), where $\tilde{v}_7$ has been moved to $v_7$, and the connection between $v_4$ and $v_7$ is removed due to expired age.}
\label{fig:ng}
\end{figure*}

Noting that some variants of NG~\cite{GNG,IGNG} use different approaches to construct NG incrementally. To be consistent with FSCIL, we directly modify the original version~\cite{NG} and learn a fixed set of nodes for the base classes.
As NG~\cite{NG} is originally learnt from unlabelled data, to accomplish the supervised incremental learning, we redefine the NG node $j$ as a tuple $v_j = (\vm_j, \mathbf{\Lambda}_j, \vz_j, c_j) \in V$, where $\vm_j \in \mathbb{R}^n$ is the centroid vector representing $\calF_j$, the diagonal matrix $\mathbf{\Lambda}_j \in \mathbb{R}^{n\times n}$ stores the variance of each dimension of $\vm_j$, and $\vz_j$ and $c_j$ are the assigned images and labels for computing the observation $\hat{\vm}_j$. With $c_j$, we can determine whether $v_j$ corresponds to old class or new class.

At the initial session $(t=1)$, the NG net with $N^{(1)}$ nodes $G^{(1)}=\langle V^{(1)}, E^{(1)}\rangle$ is trained on the feature set $\calF^{(1)}=\lbrace f(\vx; \theta^{(1)}) \vert \forall \vx \in D^{(1)} \rbrace$ using \emph{competitive Hebbian learning}. Concretely, given an input $\vf \in \calF^{(1)}$, its distance with each NG node is computed and stored in $\calD_{\vf} = \lbrace d(\vf, \vm_i) \vert i=1,\cdots,N^{(1)}\rbrace$. $\calD_{\vf}$ is then sorted in ascending order to get the rank of the nodes 
$\calR_{\vf} = \lbrace r_i | d(\vf, \vm_{r_i}) \leq d(\vf, \vm_{r_{i+1}}), \ i=1,\cdots,N^{(1)}-1 \rbrace$.
Then, for each node $r_i$, its centroid $\vm_{r_i}$ is updated to $\vm_{r_i}^*$:
\begin{equation}\label{eq:ng_update}
\vm_{r_i}^* = \vm_{r_i} + \eta \cdot e^{-i/\alpha}(\vf - \vm_{r_i}), \ i=1,\cdots,N^{(1)}-1,
\end{equation}
where $\eta$ is the learning rate, and $e^{-i/\alpha}$ is a decay function controlled by $\alpha$. We use the superscript $^*$ to denote the updated one. For the nodes distant from $\vf$, they are less affected by the update. 
Next, the edge of all connections of $r_1$ is updated as:
\begin{equation}\label{eq:topo_update}
a_{r_1 j}^* = \begin{cases}
  1 , & j = r_2; \\
  a_{r_1 j} + 1 , & j \neq r_2.
  \end{cases},
e_{r_1 j}^* = \begin{cases}
  1 , & j = r_2; \\
  0 , & a_{r_1 j}^* > T; \\
  e_{r_1 j} , & \text{otherwise}.
  \end{cases}.
\end{equation}
Apparently, $r_1$ and $r_2$ are the nearest and the second nearest to $\vf$. Their edge $e_{r_1 r_2}$ and the corresponding age $a_{r_1 j}$ is set to 1 to create or maintain a connection between node $r_1$ and $r_2$. For other edges, if $a_{r_1 j}$ exceeds lifetime $T$, the connection is removed by setting $e_{r_1 j}=0$. After training on $\calF^{(1)}$, for $v_j = (\vm_j, \vLambda_j, \vz_j, c_j)$, we pick the sample from $D^{(1)}$ whose feature vector $\vf$ is the nearest $\vm_j$ as the pseudo image $\vz_j$ and label $c_j$. The variance $\vLambda_j$ is estimated using the feature vectors whose winner is $j$.

At the incremental session $(t>1)$, for $K$-shot new class training samples, we grow $G^{(t)}$ by inserting $k < K$ (e.g. $k=1$ for $K=5$) new nodes $\lbrace \tilde{v}_N, \cdots, \tilde{v}_{N+k} \rbrace$ for each new class, and update their centroids and edges using Eq.~\eqref{eq:ng_update} and~\eqref{eq:topo_update}.  
To avoid forgetting old class, we stabilize the subgraph of NG learned at previous session $(t-1)$ that preserves old knowledge. On the other hand, to prevent overfitting to $D^{(t)}$, we enhance the discriminative power of the learned features by adapting newly inserted NG nodes and edges. The neural gas stabilization and adaptation are described in the following sections.

\subsection{Less-Forgetting Neural Gas Stabilization}

Given NG $G^{(t)}$, we extract the subgraph $G_o^{(t)} = \langle V_o^{(t)}, E_o^{(t)} \rangle \subseteq G^{(t)}$ whose vertices $v=(\vm, \vLambda, \vz, c)$ were learned on old class training data at session $(t-1)$, where $c \in \mathop{\cup}_{i=1}^{t-1} L^{(i)}$.
During finetuning, we stabilize $G_o^{(t)}$ to avoid forgetting the old knowledge. This is implemented by penalizing the shift of $v$ in the feature space $\calF^{(t)}$ via constraining the observed value of the centroid $\hat{\vm}$ to stay close to the original one $\vm$. It is noteworthy that some dimensions of $\vm$ have high diversity with large variance. These dimensions may encode common semantic attributes shared by both the old and new classes. Strictly constraining them may prevent positive transfer of the knowledge and bring unsatisfactory trade-off. 
Therefore, we measure each dimension's importance for old class knowledge using the inverted diagonal $\vLambda^{-1}$, and relax the stabilization of high-variance dimensions.
We define the \emph{anchor loss} (AL) term for less-forgetting stabilization:
\begin{align}\label{eq:al}
\ell_{AL}(G^{(t)};\theta^{(t)}) &=  \sum_{(\vm, \vLambda, \vz, c) \in V_o^{(t)} } (\hat{\vm} - \vm)^\top \vLambda^{-1} (\hat{\vm} - \vm), \nonumber \\
\text{where} \ \ \hat{\vm} &= f(\vz; \theta^{(t)}).
\end{align}
The effect of AL term is illustrated in Figure~\ref{fig:ng} (a-c). It avoids severe distortion of the feature space topology.

\subsection{Less-Overfitting Neural Gas Adaptation}

Given the new class training set $D^{(t)}$ and NG $G^{(t)}$, for a training sample $(\vx, y) \in D^{(t)}$, we extract its feature vector $\vf = f(\vx; \theta^{(t)})$ and feed $\vf$ to the NG. We hope $\vf$ matches the node $v_j$  whose label $c_j = y$, and $d(\vf, \vm_j) \ll d(\vf, \vm_i), \ i\neq j$, so that $\vx$ is more probable to be correctly classified. However, simply finetuning on the small training set $D^{(t)}$ could cause severe overfitting, where the test sample with ground-truth label $y$ is very likely to activate the neighbor with a different label. To address this problem, a \emph{min-max loss} (MML) term is introduced to constrain $\vf$ and the centroid vector $\vm_j$ of $v_j$.  The ``min'' term minimizes $d(\vf, \vm_j)$. The ``max'' term maximizes $d(\vm_i, \vm_j)$ to be larger than a margin, where $\vm_i$ is the centroid vectors of $v_j$'s neighbors with a different label $c_i \neq y$. MML is defined as: 
\begin{align}\label{eq:mml}
& \ell_{MML}(D^{(t)},G^{(t)};\theta^{(t)}) =
  \sum_{\forall (\vx,y), c_j=y} d(f(\vx; \theta^{(t)}), \vm_j) - \nonumber \\
  &\sum_{c_i\neq y, e_{ij}=1} \text{min}(0, d(\vm_i,\vm_j) - \xi).
\end{align}
The hyper-parameter $\xi$ is used to determine the minimum distance. If $d(\vm_i, \vm_j) > \xi$, we regard the distance is larger enough for well separation, and disable the term. Heuristically, we set $\xi \approx \text{max} \lbrace d(\vm_i, \vm_j) \vert \forall i,j \rbrace$. After finetuning, we update the edge $e_{ij}$ according to Eq.~\eqref{eq:topo_update}, as illustrated in Figure~\ref{fig:ng} (e) and (f). 

\subsection{Optimization}

At the incremental session $t>1$, we finetune CNN $\Theta^{(t)}$ on $D^{(t)}$ with mini-batch SGD. Meanwhile, we update the NG net $G^{(t)}$ at each SGD iteration, using the competitive learning rules in Eq.~\eqref{eq:ng_update} and~\eqref{eq:topo_update}. The gradients in Eq.~\eqref{eq:al} and~\eqref{eq:mml} are computed and back-propagated to CNN's feature extractor $f(\cdot;\theta^{(t)})$.
%To facilitate the less-forgetting AL term, we also use the images and labels assigned to $G^{(t)}$'s vertices for computing the cross-entropy loss, which is $Z^{(t)} = \lbrace (\vz, c) \vert (\vm, \vLambda, \vz, c) \in V^{(t)} \rbrace$.
The overall loss function at session $t$ is defined as:
\begin{align}\label{eq:all}
& \ell(D^{(t)}, G^{(t)};\Theta^{(t)}) =  \sum_{(\vx,y) \in D^{(t)}} -\log \hat{p}_y(\vx) + \nonumber \\
& \lambda_1\ell_{AL}(G^{(t)};\theta^{(t)}) + \lambda_2\ell_{MML}(D^{(t)},G^{(t)};\theta^{(t)}),
\end{align}
where the first term in the right-hand side is the softmax cross-entropy loss, $\ell_{AL}$ is the AL term defined in Eq.~\eqref{eq:al}, $\ell_{MML}$ is the MML term defined in Eq.~\eqref{eq:mml}, and $\lambda_1$ and $\lambda_2$ are the hyper-parameters to balance the strength.

\section{Experiment}

\begin{figure*}[h]
\begin{center}
	\includegraphics[width=1\textwidth]{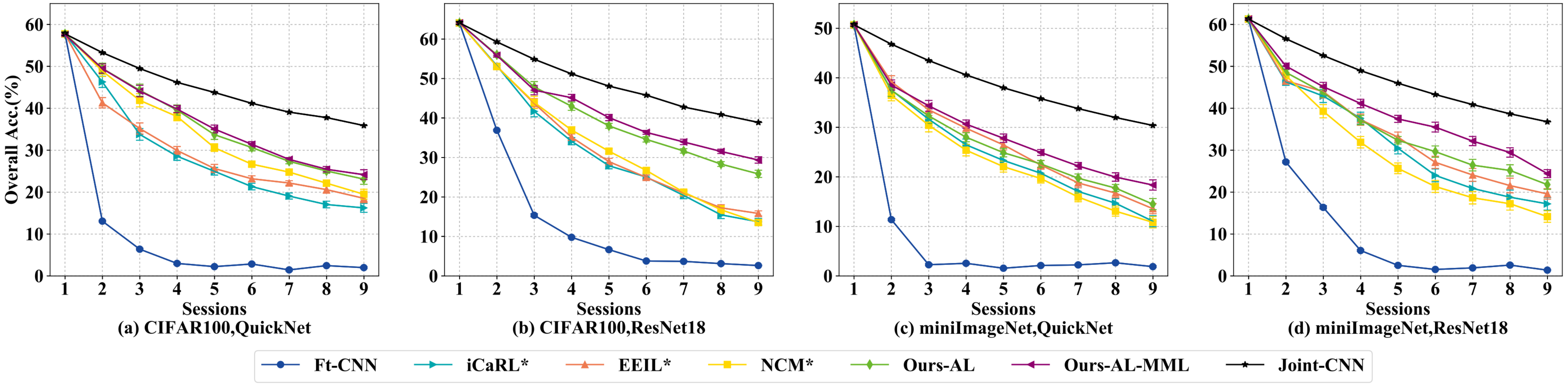}
\end{center}
\caption{Comparison of the test accuracies of QuickNet and ResNet18 on CIFAR100 and miniImageNet dataset. At each session, the models are evaluated on a joint set of test samples of the classes encountered so far.}
\label{fig:results}
\end{figure*}

We conduct comprehensive experiments on three popular image classification datasets CIFAR100~\cite{cifar}, miniImageNet~\cite{miniImageNet} and CUB200~\cite{cub200}.

\noindent\textbf{CIFAR100} dataset contains 60,000 RGB images of 100 classes, where each class has 500 training images and 100 test images. Each image has the size $32\times 32$. This dataset is very popular in CIL works~\cite{ICARL,EEIL}.

\noindent\textbf{MiniImageNet} dataset is the 100-class subset of the ImageNet-1k \cite{ImageNet} dataset used by \emph{few-shot learning}~\cite{miniImageNet,MAML}. Each class contains 500 training images and 100 test images. The images are in RGB format of the size $84\times 84$. 

\noindent\textbf{CUB200} dataset is originally designed for fine-grained image classification and introduced by \cite{AGEM,review} for incremental learning. It contains about 6,000 training images and 6,000 test images over 200 bird categories. The images are resized to $256\times 256$ and then cropped to $224\times 224$ for training.  

For CIFAR100 and miniImageNet datasets, we choose 60 and 40 classes as the base and new classes, respectively, and adopt the \emph{5-way 5-shot} setting, which we have 9 training sessions (i.e., 1 base + 8 new) in total. While for CUB200, differently, we adopt the \emph{10-way 5-shot} setting, by choosing 100 classes as the base classes and splitting the remaining 100 classes into 10 new class sessions. For all datasets, each session's training set is constructed by randomly picking 5 training samples per class from the original dataset, while the test set remains to be the original one, which is large enough to evaluate the generalization performance for preventing overfitting.

We use a shallower QuickNet~\cite{Caffe} and the deeper ResNet18~\cite{ResNet} models as the baseline CNNs. The QuickNet is a simple yet power CNN for classifying small images, which has three conv layers and two fc layers, as shown in Table~\ref{tb:quicknet}. We evaluate it on both CIFAR100 and miniImageNet. While for ResNet18, we evaluate it on all the three datasets. We train the base model $\Theta^{(1)}$ with a mini-batch size of 128 and the initial learning rate of 0.1. We decrease the learning rate to \XY{0.01} and \XY{0.001} after \XY{30} and \XY{40} epochs, respectively, and stop training at epoch \XY{50}. Then, we finetune the model $\Theta^{(t)}$ on each subsequent training set $D^{(t)}, t>1$ for \XY{100} epochs, with a learning rate of \XY{0.1} (and \XY{0.01} for CUB200). As $D^{(t)}$ contains very few training samples, we use all of them to construct the mini-batch for incremental learning. After training on $D^{(t)}$, we test $\Theta^{(t)}$ on the union of the test sets of all encountered classes. For data augmentation, we perform standard random cropping and flipping as in \cite{ResNet,NCM} for all methods. When finetuning ResNet18, as we only have very few new class training samples , it would be problematic to compute batchnorm. Thus, we use the batchnorm statistics computed on $D^{(1)}$ and fix the batchnorm layers during finetuning. We run the whole learning process \XY{10} times with different random seeds and report the average test accuracy over all encountered classes. 

\begin{table}[htb!]
\centering
\caption{The structure of the QuickNet model in the experiments, which is originally defined in the Caffe package~\cite{Caffe}.}
%\resizebox{\textwidth}{22mm}{
\begin{tabular}{lccccc}
\hline
Name & layer type & filters & filter size & stride & pad \\
\hline
conv1 & conv & 32 & 5 & 1 & 2 \\
pool1 & max pool & - & 3 & 2 & 0 \\
relu1 & relu & - & - & - & - \\
conv2 & conv & 32 & 5 & 1 & 2 \\
relu2 & relu & - & - & - & - \\
pool2 & \CXY{ave pool} & - & 3 & 2 & 0 \\
conv3 & conv & 64 & 5 & 1 & 2 \\
relu3 & relu & - & - & - & - \\
pool3 & ave pool & - & 3 & 2 & 0 \\
fc1 & fc & 64 & - & - & - \\
fc2 & fc & 100 & - & - & - \\
\hline
\end{tabular}
%}
\label{tb:quicknet}
\end{table}

We learn a NG net of \XY{400} nodes for base classes, and incrementally grow it by inserting \XY{1} node for each new class. For the hyper-parameters, we set \XY{$\eta=0.02, \alpha=1$} for faster learning of NG in Eq.~\eqref{eq:ng_update}, the lifetime \XY{$T=200$} in Eq.~\eqref{eq:topo_update}, and \XY{$\lambda_1 = 0.5, \lambda_2 = 0.005$} for Eq.~\eqref{eq:all}. 

For comparative experiments, we run the representative CIL methods in our FSCIL setting, including the classical iCARL \cite{ICARL} and the state-of-the-art methods EEIL~\cite{EEIL} and NCM~\cite{NCM}, and compare our method with them. %We have found their original training techniques are not suitable for the few-shot settings. To get reasonable performance, we use the same training settings (e.g., learning rate) for these methods as ours. 
While for BiC \cite{BIC}, we found that training the bias-correction model requires a large set of validation samples, which is impracticable for FSCIL. Therefore, we do not eval this work. We set \XY{$\gamma = 1$} in Eq.~\eqref{eq:dist} for these distillation-based methods as well as the distillation term used in our ablation study in Section~\ref{sec:ablat}. 
%Moreover, we have also compared with recent works in Incremental FSL, including Dynamic FSL~\cite{DFS} and Incremental FSL~\cite{IFS}. These two methods only support one incremental stage, since their original task aims to combine the base and novel predictor for FSL, which is different from our FSCIL.  
Other related works \cite{LWF,EWC,SI,IMM,GEM} are designed for the MT setting, which we do not involve in our experiments. We use the abbreviation ``Ours-AL'', ``Ours-AL-MML'' to indicate the applied loss terms during incremental learning. 

\begin{table*}[htb!]
\renewcommand\arraystretch{1.3}
\footnotesize
\centering
\caption{Comparison results on CUB200 with ResNet18 using the \emph{10-way 5-shot} FSCIL setting. Noting that the comparative methods with their original learning rate settings have much worse test accuracies on CUB200. We carefully tune their learning rates and boost their original accuracies by 2\%$\sim$8.7\%. In the table below, we report their accuracies after the improvement.}
%\resizebox{\textwidth}{22mm}{
\begin{tabular}{lcccccccccccc}
\hline
\multirow{2}{*}{Method} & \multicolumn{11}{c}{sessions} & our relative \\
  \cline{2-12} & 1 & 2 & 3 & 4 & 5 & 6 & 7 & 8 & 9 & 10 & 11 & improvements \\
  \hline

Ft-CNN & 68.68 & 44.81 & 32.26 & 25.83 & 25.62 & 25.22 & 20.84 & 16.77 &  18.82 & 18.25 &  17.18 & \textbf{+9.10} \\
Joint-CNN & 68.68 & 62.43 & 57.23 & 52.80 & 49.50 & 46.10 & 42.80 & 40.10 & 38.70 & 37.10 & 35.60 & upper bound \\ \hline
iCaRL*~\cite{ICARL} & 68.68 & 52.65 & 48.61 & 44.16 & 36.62 & 29.52 & 27.83 & 26.26 & 24.01 & 23.89 & 21.16 & \textbf{+5.12} \\
EEIL*~\cite{EEIL} & 68.68 & 53.63 & 47.91 & 44.20 & 36.30 & 27.46 & 25.93 & 24.70 & 23.95 & 24.13 & 22.11 & \textbf{+4.17} \\
NCM*~\cite{NCM} & 68.68 & 57.12 & 44.21 & 28.78 & 26.71 & 25.66 & 24.62 & 21.52 & 20.12 & 20.06 & 19.87 & \textbf{+6.41} \\ \hline
\textbf{Ours-AL} & \textbf{68.68} & \textbf{61.01} & \textbf{55.35} & \textbf{50.01} & \textbf{42.42} & \textbf{39.07} & \textbf{35.47} & \textbf{32.87} & \textbf{30.04} & \textbf{25.91} & \textbf{24.85} & \textbf{+1.43}\\
\textbf{Ours-AL-MML} & \textbf{68.68} & \textbf{62.49} & \textbf{54.81} & \textbf{49.99} & \textbf{45.25} & \textbf{41.40} & \textbf{38.35} & \textbf{35.36} & \textbf{32.22} & \textbf{28.31} & \textbf{26.28} & \\
\hline

% Ft-CNN & 68.68 & 44.81 & 25.05 & 17.72 & 18.08 & 16.95 & 15.10 & 10.60 & 8.93 & 8.93 & 8.47 & \textbf{+17.81} \\
% Joint-CNN & 68.68 & 62.43 & 57.23 & 52.80 & 49.50 & 46.10 & 42.80 & 40.10 & 38.70 & 37.10 & 35.60 & upper bound \\ \hline
% iCaRL* \cite{ICARL} & 68.68 & 60.50 & 46.19 & 31.87 & 29.07 & 21.86 & 21.22 & 19.15 & 16.50 & 14.46 & 14.14 & \textbf{+12.14} \\
% EEIL* \cite{EEIL} & 68.68 & 57.64 & 42.91 & 28.16 & 27.05 & 25.52 & 25.08 & 22.06 & 19.93 & 19.74 & 19.61 & \textbf{+6.67} \\
% NCM* \cite{NCM} & 68.68 & 62.55 & 50.33 & 45.07 & 38.25 & 32.58 & 28.71 & 26.28 & 23.80 & 19.91 & 17.82 & \textbf{+8.46} \\ \hline
% \textbf{Ours-AL} & \textbf{68.68} & \textbf{61.01} & \textbf{55.35} & \textbf{50.01} & \textbf{42.42} & \textbf{39.07} & \textbf{35.47} & \textbf{32.87} & \textbf{30.04} & \textbf{25.91} & \textbf{24.85} & \textbf{+1.43}\\
% \textbf{Ours-AL-MML} & \textbf{68.68} & \textbf{62.49} & \textbf{54.81} & \textbf{49.99} & \textbf{45.25} & \textbf{41.40} & \textbf{38.35} & \textbf{35.36} & \textbf{32.22} & \textbf{28.31} & \textbf{26.28} & \\
% \hline

\end{tabular}
%}
\label{tb:cub200}
\end{table*}

\subsection{Comparative results}

We report the comparative results of the methods using the \emph{5/10-way 5-shot} FSCIL setting. As the \emph{5-shot} training samples are randomly picked, we run all methods for \XY{10} times and report the average accuracies.
%\footnote{Please refer to Supplementary Material for detailed results.}
Figure~\ref{fig:results} compares the test accuracies on CIFAR100 and miniImageNet dataset, respectively. Table~\ref{tb:cub200} reports the test accuracies on CUB200 dataset.

%We show the changes of accuracies on old and new classes during incremental learning to compare the models' abilities of both ``learning new'' and ``remembering old'', respectively. 
We summarize the results as follows:
\begin{itemize}
\item On three datasets, and for both QuickNet and ResNet18 models, our TOPIC outperforms other state-of-the-art methods on each encountered session, and is the closest to the upper bound ``Joint-CNN'' method. As the incremental learning proceeds, the superiority of TOPIC becomes more significant, demonstrating its power for continuously learning longer sequence of new class datasets.
\item Simply finetuning with few training samples of new classes (i.e., ``Ft-CNN'', the blue line) deteriorates the test accuracies drastically due to \emph{catastrophic forgetting}. Finetuning with AL term (i.e., the green line) effectively alleviates forgetting, outperforming the na\"ive finetuning approach by up to \XY{\textbf{38.90\%}}. Moreover, using both AL and MML terms further achieves up to \XY{\textbf{5.85\%}} accuracy gain than using AL alone. It shows that solving the challenging FSCIL problem requires both alleviating the forgetting of the old classes and enhancing the representation learning of the new classes.
\item On CIFAR100, TOPIC achieves the final accuracies of \CXY{\textbf{24.17\%}} and \CXY{\textbf{29.37\%}} with QuickNet and ResNet18, respectively, while the second best ones (i.e., NCM$^*$ and EEIL$^*$) achieve the accuracies of \CXY{19.50\%} and \CXY{15.85\%}, respectively. TOPIC outperforms the two state-of-the-art methods by up to \CXY{\textbf{13.52\%}}. 
\item On miniImageNet, TOPIC achieves the final accuracies of \CXY{\textbf{18.36\%}} and \CXY{\textbf{24.42\%}} with QuickNet and ResNet18, respectively, while the corresponding accuracies achieved by the second best EEIL$^*$ are \CXY{13.59\%} and \CXY{19.58\%}, respectively. TOPIC outperforms EEIL* by up to \CXY{\textbf{4.84\%}}. 
\item On CUB200, at the end of the entire learning process, TOPIC achieves the accuracy of \XY{\textbf{26.28\%}} with ResNet18, outperforming the second best EEIL$^*$ (\CXY{22.11\%}) by up to \CXY{\textbf{4.17\%}}. 
\end{itemize}

\begin{table*}[htb!]
\renewcommand\arraystretch{1.3}
\footnotesize
\centering
\caption{Comparison results of combining different loss terms on miniImageNet with ResNet18.}
%\resizebox{\textwidth}{24mm}{
\begin{tabular}{lccccccccccccc}
\hline
\multirow{2}{*}{Method} & \multirow{2}{*}{DL} & \multirow{2}{*}{AL} & \multirow{2}{*}{min term} & \multirow{2}{*}{max term} & \multicolumn{9}{c}{sessions} \\
  \cline{6-14} & & & & & 1 & 2 & 3 & 4 & 5 & 6 & 7 & 8 & 9   \\
  \hline

baseline DL & \checkmark & & & & 61.31 & 46.85 & 42.34 & 36.56 & 30.63 & 27.64 & 24.61 & 22.06 & 18.69 \\
DL-MML & \checkmark & & \checkmark & \checkmark & 61.31 & 48.14 & 42.83 & 38.35 & 32.76 & 30.02 & 27.70 & 25.43 & 20.55 \\ \hline
baseline AL & & \checkmark & & & 61.31 & 48.58 & 43.77 & 37.19 & 32.38 & 29.67 & 26.44 & 25.18 & 21.80 \\
AL w/o. $\bf{\Lambda}$ & & \checkmark & &  & 61.31 & 48.55 & 42.73 & 36.73 & 32.59 & 28.40 & 25.23 & 23.69 & 21.36 \\
AL-Min & & \checkmark & \checkmark & & 61.31 & \textbf{50.60} & 45.14 & 41.03 & 35.69 & 33.64 & 30.11 & 27.79 & 24.18 \\
AL-Max & & \checkmark &  & \checkmark & 61.31 & 48.49 & 43.03 & 38.53 & 34.24 & 31.79 & 28.96 & 26.09 & 23.80 \\
AL-MML & & \checkmark & \checkmark & \checkmark & 61.31 & 50.09 & \textbf{45.17} & \textbf{41.16} & \textbf{37.48} & \textbf{35.52} & \textbf{32.19} & \textbf{29.46} & \textbf{24.42} \\
AL-MML w. DL & \checkmark & \checkmark & \checkmark & \checkmark & 61.31 & 50.00 & 44.23 & 39.85 & 36.02 & 32.95 & 29.78 & 27.17 & 23.49 \\
\hline

\end{tabular}
%}
\label{tb:ablat_loss}
\end{table*}

\begin{table}[htb!]
\renewcommand\arraystretch{1.3}
\footnotesize
\centering
\caption{Comparison of the final test accuracies achieved by ``exemplars'' and NG nodes with different memory size. Experiments are performed on CIFAR100 with ResNet18.}
\begin{center}
\begin{tabular}{lccccccc}

  \hline
  Memory  & 50 & 100 & 200 & 400 & 800 & 1600  \\
  \hline
  Exemplars & 19.21 & 22.32 & 26.94 & 28.25 & 28.69 & 28.89 \\
  NG nodes & 22.37 & 25.72 & 28.56 & 29.37 & 29.54 & 29.35 \\
  \hline

\end{tabular}
\end{center}
\label{tb:exemplar}
\end{table}

\begin{figure}[htb!]
\begin{center}
    \includegraphics[width=0.48\textwidth]{figures/overall3.pdf}
\end{center}
\caption{Comparison results under the \emph{5-way 10-shot} and \emph{5-way full-shot} settings, evaluated with ResNet18 on miniImageNet.}
\label{fig:shots}
\end{figure}

\subsection{Ablation study}\label{sec:ablat}

\noindent\textbf{The contribution of the loss terms.} We conduct ablation studies to investigate the contribution of the loss terms to the final performance gain. The experiments are performed on miniImageNet with ResNet18.  For AL, we compare the original form in Eq.~\eqref{eq:al} and a simplified form without the ``re-weighting'' matrix $\bf{\Lambda}$. For MML, as it consists of the ``min'' and ``max'' terms, we evaluate the performance gain brought by each term separately. Besides, we also investigate the impact brought by the distillation loss term, which is denoted as ``DL''.
Table~\ref{tb:ablat_loss} reports the comparison results of different loss term settings. We summarize the results as follows:
\begin{itemize}
\item The ``AL'' term achieves better accuracy (up to \textbf{1.49\%}) than the simplified form ``AL w/o. $\vLambda$'', thanks to the feature re-weighting technique.
\item Both ``AL-Min'' and ``AL-Max'' improve the performance of AL, and the combined form ``AL-MML'' achieves the best accuracy, exceeding ``AL'' by up to \textbf{5.85\%}.
\item Both ``DL-MML'' and ``AL-MML'' improve the performance of the corresponding settings without MML (i.e., ``DL'' and ``AL''). It demonstrate the effectiveness of the MML term for improving the representation learning for few-shot new classes.
\item Applying the distillation loss degrades the performance. Though distillation is popularly used by CIL methods, it may be not so effective for FSCIL, as it is difficult to balance the old and new classes and trade-off the performance when there are only few new class training samples, as discussed in Section~\ref{sec:baseline}.
%\footnote{Please refer to Supplementary Material Figure 7 for the comparison of the confusion matrix produced by distillation-based methods.}    
\end{itemize}

\noindent\textbf{Comparison between ``exemplars'' and NG nodes}.
In our method, we represent the knowledge learned in CNN's feature space using the NG net $G$. An alternative approach is to randomly select a set of exemplars representative of the old class training samples \cite{ICARL,EEIL} and penalize the changing of their feature vectors during training.  Table~\ref{tb:exemplar} compares the final test accuracies achieved by the two approaches under different memory sizes.  From Table~\ref{tb:exemplar}, we can observe that \emph{using NG with only a few number of nodes can greatly outperform the exemplar approach in a consistent manner}. When smaller memory is used, the difference in accuracy becomes larger, demonstrating the superiority of our method for FSCIL.

\noindent\textbf{The effect of the number of training samples.} To investigate the effect brought by different shot of training samples, we further evaluate the methods under the \emph{5-way 10-shot} and \emph{5-way full-shot} settings. For \emph{5-way full-shot}, we use all training samples of the new class data, which is analogous to the ordinary CIL setting. We grow NG by adding 20 nodes for each new session, which we have $(400+20(t-1))$ NG nodes at session $(t-1)$.
Figure~\ref{fig:shots} shows the comparative results of different methods under the \emph{10-shot} and \emph{full-shot} settings. We can see that our method also outperforms other state-of-the-art methods when training with more samples. It demonstrate the effectiveness of the proposed framework for general CIL problem. 

\begin{figure*}[t]
\setcounter{figure}{5}
\begin{center}
	\includegraphics[width=1\textwidth]{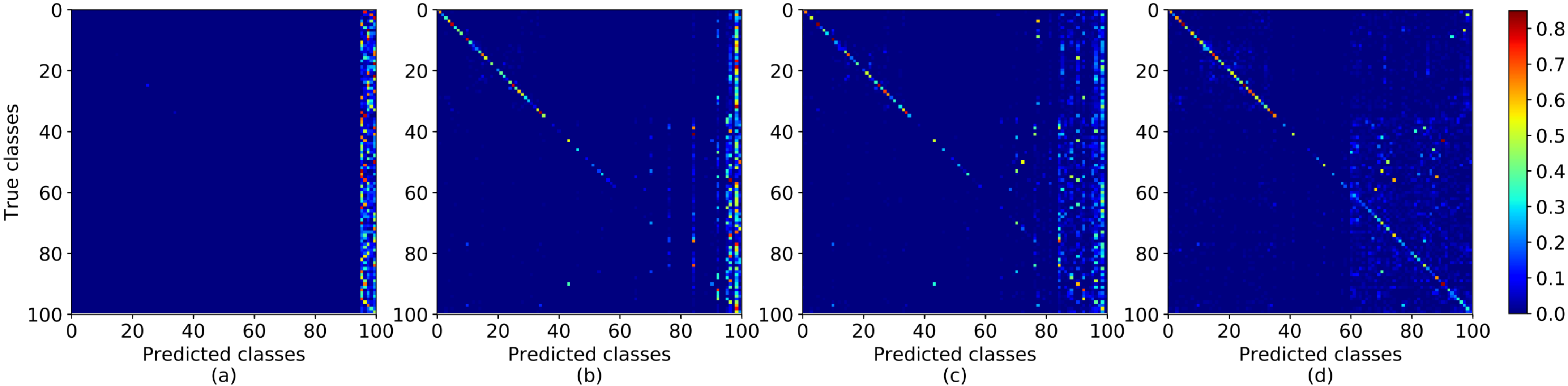}
\end{center}
\caption{Comparison of the confusion matrices produced by (a) Ft-CNN, (b) EEIL*, (c) NCM*, and (d) our TOPIC on miniImageNet with ResNet18.}
\label{fig:confusion}
\end{figure*}

Figure~\ref{fig:confusion} compares the confusion matrix of the classification results at the last session, produced by Ft-CNN, EEIL*~\cite{EEIL}, NCM*~\cite{NCM} and our TOPIC. 
The na\"ive finetuning approach tends to misclassify all past classes (i.e., 0-94) to the newly learned classes (i.e., 95-99), indicating \emph{catastrophic forgetting}. EEIL* and NCM* can alleviate forgetting to some extent, while still tend to misclassify old class test samples as new classes due to overfitting. Our method, named ``TOPIC'', produces a much better confusion matrix, where the activations are mainly distributed at the diagonal line, indicating higher recognition performance over all encounter class.  It demonstrate the effectiveness of solving FSCIL by avoiding both ``forgetting old'' and ``overfitting new''.

\section{Conclusion}

We focus on a unsolved, challenging, yet practical incremental-learning scenario, namely the \emph{few-shot class-incremental learning} (FSCIL) setting, where models are required to learn new classes from few training samples. We propose a framework, named TOPIC, to preserve the knowledge contained in CNN's feature space. TOPIC uses a \emph{neural gas} (NG) network to maintain the topological structure of the feature manifold formed by different classes. We design mechanisms for TOPIC to mitigate the forgetting of the old classes and improve the representation learning for few-shot new classes. Extensive experiments show that our method substantially outperforms other state-of-the-art CIL methods on CIFAR100, miniImageNet, and CUB200 datasets, with a negligibly small memory overhead.

%We focus on a unsolved, challenging, yet practical incremental-learning scenario, as defined as the \emph{few-shot class-incremental learning} (FSCIL) setting, where models are required to learn new classes from few training samples. We propose a framework, named TOPIC, to preserve the knowledge contained in CNN's feature space for FSCIL. TOPIC uses a \emph{neural gas} (NG) network to preserve the topological structure of the feature manifold formed by different classes. We design mechanisms for TOPIC to mitigate the forgetting of the old classes and improve the representation learning for few-shot new classes. Extensive experiments show that our method significantly outperforms other state-of-the-art CIL methods on CIFAR100, miniImageNet and CUB200 datasets, with a negligibly small memory overhead.

%\noindent\textbf{Acknowledgements.} The authors express deep gratitude to National Key R\&D Program of China under Grand No.2019YFB1312000 and National Major
%Project under Grant No.2017YFC0803905. 

{\small
\bibliographystyle{ieee_fullname}
\bibliography{main}
}

\end{document}